\documentclass[journal,twoside,web]{ieeecolor}
\usepackage{generic}
\usepackage{cite}
\usepackage{amsmath,amssymb,amsfonts}
\usepackage{algorithmic}
\usepackage{graphicx}
\usepackage{algorithm,algorithmic}
\usepackage{hyperref}
\hypersetup{hidelinks=true}
\usepackage{textcomp}
\usepackage{booktabs} 
\usepackage{multirow}
\usepackage{array}
\def\BibTeX{{\rm B\kern-.05em{\sc i\kern-.025em b}\kern-.08em
    T\kern-.1667em\lower.7ex\hbox{E}\kern-.125emX}}
\begin{document}
\title{Uncertainty as a Proxy for Semantic Correctness in Diffusion-Based Medical
Image Synthesis}
\author{Yuxuan Ou, Konstantinos Kamnitsas, OxAAA Study, AICT Consortium,
Regent Lee, Vicente Grau
\thanks{Y. Ou, K. Kamnitsas, and V. Grau are with the Department of
Engineering Science, University of Oxford, Oxford, United Kingdom.
R. Lee is with the Nuffield Department of Surgical Sciences,
University of Oxford, Oxford, United Kingdom.}}
\maketitle

\begin{abstract}
Diffusion models can synthesise contrast-enhanced CT (CECT) from
non-contrast CT (NCCT), avoiding contrast administration and its
environmental and patient-access costs. However, visually
realistic images are not necessarily anatomically correct, and the pixel-intensity and feature-space similarity metrics
used to assess generation quality do not directly measure anatomical
correctness. In this work, we investigate whether
uncertainty can serve as a proxy for semantic correctness in
diffusion-based medical image synthesis.

We study NCCT-to-CECT synthesis using AortaDiff, a multitask diffusion
framework that jointly generates CECT images and lumen segmentations.
The segmentation output provides an explicit representation of the
generated vascular anatomy, enabling segmentation-derived errors to be
used as a quantitative measure of generation correctness. Six methods spanning weight (Ensemble, HyperDiff, BayesDiff), architecture-perturbation (MCDropout), generative-stochasticity (RDS) and input-perturbation (TTA) uncertainty are compared at the pixel, region and image levels, and for detection of clinically relevant out-of-distribution (OOD) cases.

Uncertainty proves informative at all three spatial scales, remains
informative on an external multi-centre dataset under distribution
shift, and supports OOD detection. MCDropout stands out among the six: it ranks among the leading methods
at every scale, generalizes well on the external dataset, and can be enabled at
inference on any model already trained with dropout, so reliable
uncertainty comes at no extra training cost. Uncertainty reliably flags
severe failures but discriminates poorly among already high-quality
images. These findings support uncertainty as a practical and
computationally economical signal for quality filtering, reliability
assessment and OOD detection in NCCT-to-CECT synthesis.
\end{abstract}


\section{Introduction}
\label{sec:introduction}

\begin{figure*}[t]
    \centering
    \includegraphics[width=\textwidth]{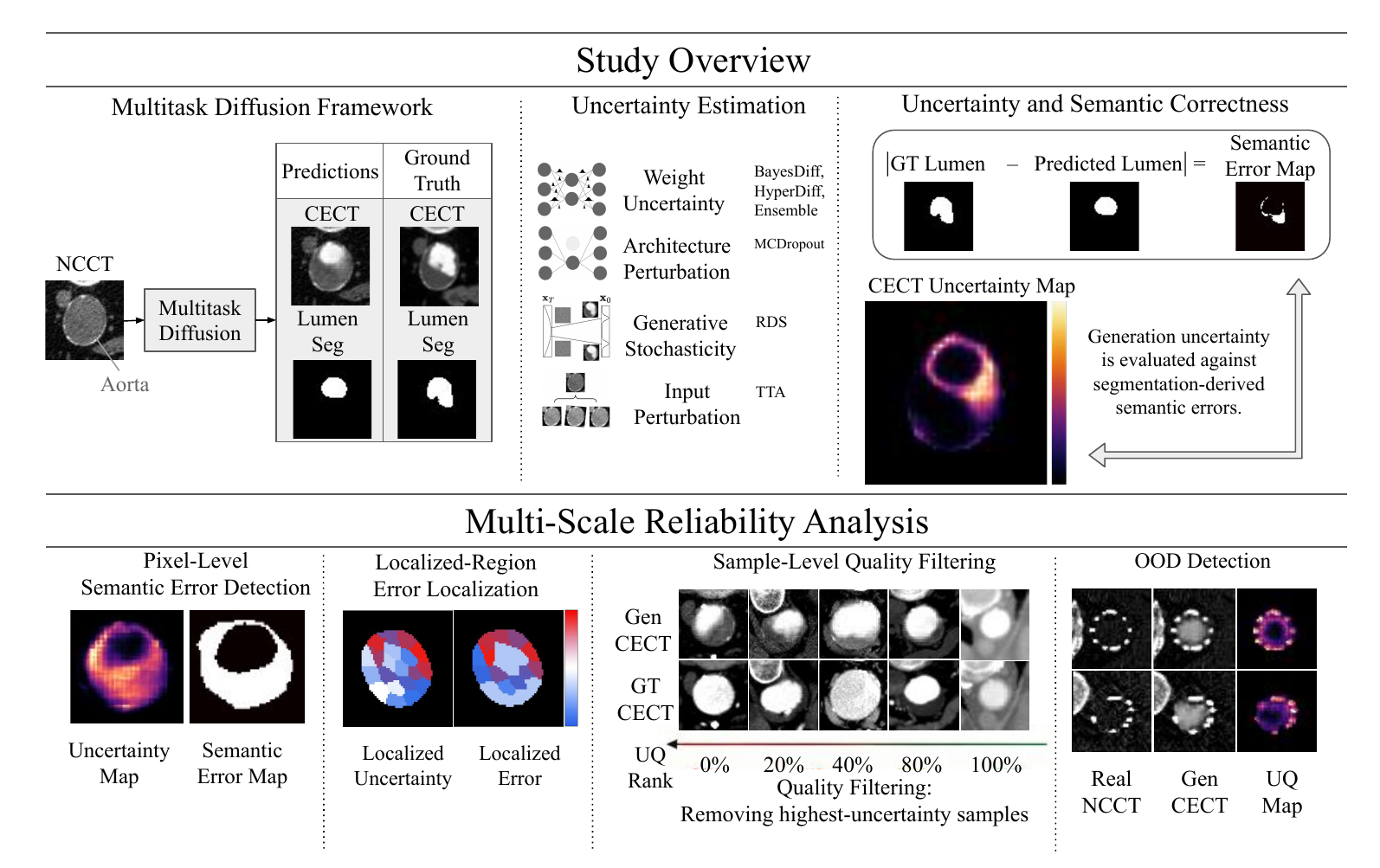}
    \caption{
    Overview of the proposed evaluation framework. AortaDiff jointly generates CECT and lumen segmentation from NCCT. Uncertainty is evaluated against segmentation-derived semantic error at pixel, region, and image levels, with additional external OOD evaluation.
    }
    \label{fig:framework}
\end{figure*}

Diffusion models have recently achieved strong performance in medical image synthesis \cite{kazerouni2023diffusion, diffusionmed2, diffusionmed1}. A clinically important application is the synthesis of contrast-enhanced images from non-contrast scans, including contrast-enhanced CT (CECT) from non-contrast CT (NCCT) \cite{aaa_dl,shiri2025generatingsyntheticcontrastenhancedchest, cect2,cect1} and contrast-enhanced from non-contrast MRI \cite{ncmri,ncmri3,ncmri2}. Contrast administration adds costs beyond the scan itself. Iodinated contrast media are poorly removed by conventional wastewater treatment and are now widely detected as environmental contaminants \cite{Sengar2021-zj}. Moreover, concerns about kidney risk may lead to contrast-enhanced studies being withheld from patients with impaired renal function \cite{Davenport2020-nv}. Synthesising CECT from NCCT therefore has implications for the environmental footprint, resource resilience, and accessibility of routine vascular imaging. 

These benefits depend on the generated vascular anatomy being trustworthy. However, a generated image that appears visually realistic is not necessarily anatomically correct \cite{NEURIPS2023_2b1d1e5a}. Small structural errors in the aorta may lead to incorrect downstream analysis and clinical decisions. Existing image quality metrics such as PSNR, SSIM, LPIPS and MSE measure similarity either between pixel
intensities or between feature representations, and do not
directly assess whether the generated anatomy is semantically correct \cite{wang2026evaluatingcalibratingdiffusionmodelderived}. Therefore, reliable assessment of generation correctness remains an important challenge \cite{KOCAK2025100030}.

Uncertainty in a diffusion model can arise from different sources, and uncertainty estimation methods differ primarily in where stochasticity or variability is introduced \cite{ABDAR2021243}. We therefore organize diffusion-based UQ methods according to four sources: weight uncertainty, which perturbs or samples the learned parameters \cite{blundell2015weight,Lakshminarayanan2016SimpleAS,kou2024bayesdiff, chan2024estimating}; architecture-perturbation uncertainty, which keeps weights fixed but stochastically modifies the computation graph, with Monte Carlo Dropout (MCDropout) as the predominant approach \cite{gal2016dropout}; generative-stochasticity uncertainty, which captures the variability of diffusion generation across repeated sampling from different initial noise realizations \cite{ho2020denoising, songdenoising}; and input-perturbation uncertainty, which probes model stability under test-time transformations of the input image \cite{ayhan2018test}. These sources correspond to the main stages at which variability can enter a sampling-based generation pipeline: the model parameters, the network computation, the diffusion sampling process, and the input condition.

Existing evaluations of uncertainty in medical image generation do not test it against semantic correctness, i.e. whether the generated anatomy matches the ground truth. Some measure error as pixel-intensity or feature-space dissimilarity between generated and reference images, using metrics such as MAE, SSIM and LPIPS~\cite{Kustner2024-vy, Kutiel2023-jl, ekmekci2025conformalizedgenerativebayesianimaging, Teneggi2026-at}, which capture anatomical error only indirectly. Others validate uncertainty against distribution-level scores such as FID under synthetically degraded inputs~\cite{11229379}. To our knowledge, no study establishes whether uncertainty corresponds to semantic errors in the generated anatomy.

We study these uncertainty quantification methods using a multitask diffusion framework for NCCT to CECT synthesis \cite{ou2025aortadiffunifiedmultitaskdiffusion}. The model jointly generates a CECT image and a lumen segmentation mask, as shown in Figure~\ref{fig:framework}. This provides a natural way to evaluate semantic correctness because the segmentation output directly reflects the generated aorta structure. We show that the multitask model's segmentation output agrees closely with three independently trained external segmentation networks (nnU-Net, UNet++ and Attention U-Net \cite{Isensee2021-yi,zhou2018unet++,oktay2018attention}) applied to the generated image, confirming that the segmentation branch faithfully reflects the generated vascular anatomy. Its agreement with the ground-truth lumen therefore provides a measure of generation correctness.

Using this framework, we systematically evaluate representative uncertainty estimation methods for diffusion models. We perform a multi-scale evaluation that studies uncertainty at the pixel, region, and image level, and assess whether uncertainty can detect clinically relevant out-of-distribution (OOD) cases, specifically vascular stents, on an external multi-center dataset. Together, these analyses reveal whether uncertainty corresponds to local semantic errors, whether it can flag failed generations, and whether it remains consistent on an external dataset and identifies OOD cases under distribution shift.

Our experiments lead to three main findings. First, uncertainty provides meaningful information about semantic correctness at the pixel, region and image levels, and is most informative for severe generation failures. This information is preserved on an external multi-centre dataset, where image-level failure detection remains effective. Second, MCDropout ranks among the leading methods at every scale. When the base model is trained with dropout, reliable uncertainty estimates can be obtained without training additional networks. Third, uncertainty maps can be aggregated into image-level scores for quality filtering and for detecting clinically relevant OOD structures such as vascular stents.

The main contributions of this work are:
\begin{itemize}
\item We propose an evaluation framework that tests diffusion model generation uncertainty against semantic correctness, measured by agreement with the ground-truth anatomy rather than image similarity, using the lumen segmentation jointly predicted by a multitask diffusion model.
\item We introduce a multi-scale evaluation protocol spanning the pixel, region and image levels, extended to external multi-centre data under distribution shift and to detection of clinically relevant OOD cases (vascular stents).
\item We compare six uncertainty quantification methods for diffusion models, spanning four sources of uncertainty.
\item We show that MCDropout remains among the leading methods at every scale and under distribution shift, requiring no additional training when the base model is trained with dropout.
\end{itemize}

\section{Related Work: Uncertainty Quantification Methods}

Uncertainty in diffusion models can arise from model parameters, input perturbations, architectural perturbations, and stochasticity in the diffusion sampling process \cite{ABDAR2021243}. We therefore evaluate representative methods from these four sources: weight uncertainty (BayesDiff, HyperDiff, and deep ensembles), input perturbation (TTA), architecture perturbation (MC Dropout), and generative stochasticity (repeated DDIM sampling).

\subsection{Weight Uncertainty}

Weight uncertainty models variability in the learned parameters and assumes that parameter disagreement propagates to uncertainty in the generated output \cite{Lakshminarayanan2016SimpleAS,kou2024bayesdiff,chan2024estimating}. We consider three approaches with complementary implementations.

\paragraph{BayesDiff}
BayesDiff \cite{kou2024bayesdiff} applies a last-layer Laplace approximation (LLLA) \cite{llla} to the diffusion noise-prediction network, fitting a Gaussian posterior around the pre-trained final layer. The resulting predictive variance is propagated through the reverse diffusion process, with the covariance term estimated by Monte Carlo sampling, yielding a pixel-wise uncertainty map.

\paragraph{HyperDiff}
HyperDiff \cite{chan2024estimating} uses a Bayesian hypernetwork \cite{krueger2017bayesian} to generate plausible weights for the diffusion denoising network. Epistemic uncertainty is obtained from variation across sampled weights, while aleatoric uncertainty captures variation within each weight configuration.

\paragraph{Deep Ensembles}
Deep ensembles \cite{Lakshminarayanan2016SimpleAS} approximate weight uncertainty by independently training multiple models and measuring disagreement among their predictions.

\subsection{Input Perturbation Uncertainty}

Input perturbation uncertainty measures model sensitivity to plausible changes in the input. Test-time augmentation (TTA) \cite{ayhan2018test} generates multiple predictions from randomly transformed inputs and estimates uncertainty from their variability. Unlike weight-based approaches, TTA keeps the model parameters fixed and probes sensitivity to input perturbations.

\subsection{Architecture Perturbation Uncertainty}

Architecture perturbation uncertainty estimates reliability by introducing stochasticity into the network computation during inference. Monte Carlo Dropout (MCDropout) \cite{gal2016dropout} keeps dropout active during inference, producing different effective subnetworks across repeated forward passes. Uncertainty is estimated from the variability of the resulting predictions.

\subsection{Generative Stochasticity}

Generative stochasticity captures variability arising from the diffusion sampling process. We use repeated deterministic DDIM sampling (RDS) \cite{songdenoising} with different initial noise realizations. Because the reverse process is deterministic given the initial noise, variability across samples reflects differences in the initial noise and provides a direct measure of generation diversity.

\section{Methodology}
\subsection{Multitask Diffusion Framework}
We study uncertainty quantification in the setting of NCCT to CECT
synthesis of the abdominal aorta. Given an NCCT slice, a multitask
diffusion model jointly predicts a synthetic CECT image and a lumen
segmentation mask, as shown in Fig.~\ref{fig:framework}
\cite{ou2025aortadiffunifiedmultitaskdiffusion}. The two tasks are
optimised simultaneously and share a common image representation, so
the segmentation output provides an explicit representation of the
lumen structure present in the generated image. This allows
segmentation-derived errors to be used as a semantic proxy for
generation correctness, rather than the appearance similarity measured
by PSNR, SSIM and LPIPS.
\subsection{Generation Uncertainty}
For all methods, uncertainty is estimated from the CECT generation
branch, yielding a pixel-wise uncertainty map over the synthetic CECT
image; the segmentation branch is used only to define the semantic
correctness target.

\subsection{Semantic Correctness Definition and Evaluation}
\label{subsec:semantic-correctness}
Semantic correctness is defined with respect to the aorta region.
All semantic error measures are computed from the lumen segmentation,
at three spatial scales: pixel, region and image. 
\subsubsection{Pixel Level}
Semantic error is the disagreement between the predicted and
ground-truth lumen masks, $s_{\text{pred}}$ and $s_{\text{gt}}$:
pixels where the masks disagree are semantically incorrect, giving a
binary error map. Pixel-wise AUROC, computed within the aortic region,
measures whether uncertainty separates correct from incorrect pixels.
For sparsification, let $\mathcal{A}$ be the aortic pixels and $u_i$
the uncertainty at pixel $i \in \mathcal{A}$. At fraction
$t \in [0,1]$ we remove the fraction $t$ of pixels with the highest
uncertainty, leaving $K(t)$, and recompute Dice over the retained
pixels only:
\begin{equation}
\label{eq:dice}
\mathrm{Dice}(t) = \frac{2\,\lvert (s_{\text{pred}} \cap s_{\text{gt}}) \cap K(t) \rvert}{\lvert s_{\text{pred}} \cap K(t) \rvert + \lvert s_{\text{gt}} \cap K(t) \rvert}.
\end{equation}
Removing an incorrect pixel drops a term from the denominator only and
raises retained Dice, whereas removing a correctly predicted lumen
pixel also drops the numerator and lowers it. An ideal uncertainty map
therefore follows the oracle curve, which removes pixels in order of
their true semantic error. We summarise each curve by the area under
the sparsification error curve (AUSE, the gap to the oracle, lower is
better) and the area under the sparsification curve (AUSC, the retained
Dice itself, higher is better). Both are computed from the mean curve
aggregated across test slices and therefore carry no standard
deviation.

\subsubsection{Region Level}
Pixel-wise analysis does not reveal whether uncertainty can spatially
localize errors. We partition the aortic region into superpixels using
SLIC \cite{item_605ab010dcfa4dcfbb55ddf0eda015ba} with compactness 20.
Regional semantic error is the proportion of incorrectly predicted
pixels in each superpixel $R_k$, and regional uncertainty its mean
pixel uncertainty $\bar{u}_k = \frac{1}{|R_k|}\sum_{i \in R_k} u_i$.
Spearman correlation between the two measures rank agreement across
regions. For sparsification, whole superpixels are discarded in order
of decreasing $\bar{u}_k$ until a fraction $t$ of aortic pixels has
been removed, and the retained Dice is recomputed as in
\eqref{eq:dice}; curves are again summarised by AUSE and AUSC.

\subsubsection{Image Level}
Semantic correctness is quantified by the lumen Dice of the whole
image, which distinguishes successful from failed generations. Since
the preceding analyses produce pixel-wise maps, a scalar uncertainty
score $U$ is required per image; we aggregate each map by its mean,
maximum, 90th percentile, and maximum local average (window size 7).
For rejection, each image receives a failure label
$y = \mathbf{1}[\,\mathrm{Dice} < \tau\,]$ and detection performance is
the AUROC of $U$ against $y$; because $\tau$ is
application-dependent, we sweep it across a range and report AUROC at
each. For sparsification, images are progressively removed in order of
decreasing $U$ and the average Dice of the retained images is
recomputed at each step, compared against the oracle obtained by
removing images in order of their true Dice.

\subsection{Out-of-Distribution Detection}

In addition to semantic correctness, we investigate whether uncertainty can identify clinically relevant out-of-distribution (OOD) cases. It evaluates whether uncertainty can detect previously unseen structures that were not present during training.

We use vascular stents as a clinically relevant OOD scenario. Stents are absent from the training dataset and therefore represent previously unseen structures. Pixel-wise uncertainty maps are aggregated into image-level uncertainty scores and used to classify stent-containing and non-stent slices. Detection performance is quantified using AUROC.

Together, the semantic correctness and OOD evaluations provide a comprehensive assessment of whether uncertainty can be used as a reliability signal for diffusion-based medical image synthesis, both for identifying semantic errors and for detecting clinically relevant distribution shifts.

\section{Datasets}

In this study, we use two datasets containing paired NCCT and CECT scans. Both datasets were acquired with full regulatory approval, which enabled secondary use of imaging data acquired as part of the routine healthcare service. The Oxford dataset (UK Health Research Authority Approval Ref: 13/SC/0250) is used for model training and in-distribution evaluation, while the AICT dataset (UK Health Research Authority Approval Ref: 22/HRA/2303) serves as an external validation dataset for evaluating the relationship between uncertainty and semantic error under distribution shift and OOD detection. The two datasets were collected from different clinical sites and imaging scanners providing a realistic setting for assessing the robustness.

\subsection{Oxford Dataset}
\label{subsec:oxford-dataset}

The Oxford dataset is underpinned by the Oxford Abdominal Aortic Aneurysm (OxAAA) Study \cite{oxaaa2,oxaaa1, Chandrashekar2020-bt}. It consists of 264 patient cases, each containing spatially registered NCCT and CECT scans. Aortic segmentation labels are available for all cases, while 65 cases additionally contain expert-annotated lumen segmentations. 

Following our previous work \cite{ou2025aortadiffunifiedmultitaskdiffusion}, the 199 cases without lumen annotations, together with 45 of the 65 lumen-annotated cases, were used for training and validation; the remaining 20 lumen-annotated cases were held out for testing. For model development and evaluation, volumetric scans were processed as independent 2D axial slices. The 20 test cases yielded a total of 1,547 paired NCCT--CECT slices.

Slices were retained only if the Dice coefficient between the
registered NCCT and CECT aorta masks exceeded 0.90, leaving 1,316
slices in the final test set.

\subsection{AICT Dataset}

The AICT repository is sourced from 10 hospitals across 3 continents (www.aict.ai). It is used as an external dataset to evaluate the generalizability of uncertainty estimation under data distribution shift. The dataset contains paired NCCT and CECT scans acquired from multiple clinical centers using different scanners and acquisition protocols.

As expert lumen annotations were not available, automated segmentation methods were used during preprocessing. First, NCCT and CECT volumes were aligned using rigid registration. Aortic segmentations were first generated using an nnU-Net model \cite{Isensee2021-yi} trained on the Oxford dataset \cite{Chandrashekar2020-bt}, and each result was manually inspected. When the nnU-Net segmentation was judged inadequate, TotalSegmentator \cite{doi:10.1148/ryai.230024} was applied as a fallback and used if it produced a satisfactory mask; cases for which neither method yielded an acceptable aortic segmentation were discarded. Lumen segmentations were generated using another nnU-Net model trained on the Oxford dataset, and each mask was manually reviewed.

Quality control was subsequently performed to ensure adequate image quality and successful preprocessing. Cases were retained only if: (i) the CECT scan exhibited sufficient vascular enhancement to allow clear visualization of the aortic lumen; (ii) the scan contained the abdominal aorta; (iii) NCCT and CECT volumes could be successfully aligned using rigid registration; (iv) at least one automated segmentation method successfully extracted the aortic mask; and (v) the Dice similarity coefficient between the registered NCCT and CECT aortic masks exceeded 0.90. In addition to the above quality-control criteria, all retained slices and the corresponding automatically generated aortic and lumen segmentation masks were manually reviewed.

After preprocessing and quality control, 61 cases were retained for analysis. The final cohort consisted of 21 cases from Opole, Poland; 18 from Leicester, UK; 17 from Heraklion, Greece; 3 from São Paulo, Brazil, and 2 from Brugge, Belgium. This yielded 5,950 slices, of which 1,575 (18 cases) contained vascular stents. Throughout this work, AICT was used exclusively for external evaluation, allowing us to assess the robustness of uncertainty estimation methods on previously unseen data distributions.

\section{Uncertainty Estimation Implementation Details}
\label{sec:uq_implementation}
Table~\ref{tab:methods} summarises the four uncertainty sources, the
six representative methods, and their configuration and cost. Two
details are not captured there. For BayesDiff, the Laplace
approximation is fitted on the 2{,}244 paired NCCT--CECT slices with
lumen masks from the Oxford training set, the covariance
$\mathrm{Cov}[x,\epsilon]$ between the current state and the predicted
noise is estimated with 100 Monte Carlo samples per step, and
generation uncertainty is the final propagated variance. HyperDiff
yields both aleatoric and epistemic estimates; their mean pixel-wise
Spearman correlation within the aortic region is $0.962 \pm 0.031$ and
they perform comparably at all evaluation levels, so we report
epistemic uncertainty throughout.

\begin{table*}[t]
\centering
\caption{Evaluated uncertainty estimation methods. All use DDIM
sampling with 10 steps; generation is the mean over samples and
uncertainty the pixel-wise sample variance. Inference time is per test
case.}
\label{tab:methods}

\begin{tabular}{llp{6.2cm}ccl}
\toprule
Source & Method & Mechanism and configuration & Passes & Time (s) & Training overhead \\
\midrule

Input perturbation
& TTA
& 400 augmented views (bounding-box expansion, rotation, isotropic scaling), inverse-warped to original coordinates
& 400 & 42 & None \\

Generative stochasticity
& RDS
& Deterministic DDIM ($\eta=0$) from 400 initial noise realizations
& 400 & 42 & None \\

Architecture perturbation
& MCDropout
& Dropout $p=0.5$ in U-Net residual blocks; 20 seeds $\times$ 20 trajectories
& 400 & 44 & None \\

\multirow{3}{*}{Weight}
& Ensemble
& 4 independently trained models $\times$ 100 samples
& 400 & 44 & 4 models \\

& HyperDiff
& Bayesian hypernetwork generating denoiser weights; 20 weight sets $\times$ 20 samples
& 400 & 55 & Hypernetwork \\

& BayesDiff
& Last-layer Laplace approximation (diagonal Hessian) on the final convolutional layer
& 1 & 0.22 & Laplace fitting \\

\bottomrule
\end{tabular}
\end{table*}

\section{Results}
We first validate the use of multitask segmentation as a semantic measure for generation correctness. We then investigate the relationship between generation uncertainty and semantic correctness at the pixel, region, and image levels. Finally, we evaluate uncertainty on external data and assess its utility for detecting clinically relevant out-of-distribution structures.
\subsection{Using Segmentation as a Semantic Proxy for Generation Correctness}
We adopt the training data and model configuration of our previous work
\cite{ou2025aortadiffunifiedmultitaskdiffusion}, evaluated on the
refined test set of Section~\ref{subsec:oxford-dataset}. Under DDIM
inference with 10 steps, the multitask framework improves on a
two-stage pipeline (a generation-only diffusion model followed by an
external nnU-Net) in PSNR (25.37 vs.\ 25.12), SSIM (0.84 vs.\ 0.80) and
LPIPS (0.06 vs.\ 0.08) at equal segmentation Dice (0.90).

The proxy additionally requires that the segmentation branch reflect
the lumen structure present in the model's jointly generated image.
We therefore applied three independently trained 2D lumen segmenters
--- nnU-Net \cite{Isensee2021-yi}, UNet++ \cite{zhou2018unet++} and
Attention U-Net \cite{oktay2018attention}, all trained on real CECT
annotations from the Oxford dataset --- to the generated CECT images.
The multitask segmentation agreed with each at a mean Dice of 0.95,
confirming that it faithfully represents the synthesized vascular
anatomy. Its disagreement with the reference lumen annotation therefore
provides a quantitative measure of semantic error.

\subsubsection{Robustness to Registration Quality}
\label{sec:registration_quality_robustness}
Disagreement between predicted and reference lumen masks could also
reflect residual NCCT-to-CECT registration error. We therefore repeated
the pixel-, region- and image-level analyses after stratifying the
OxAAA test slices by aortic-mask registration Dice (above 0.95,
$n=791$; 0.90--0.95, $n=525$). Across all six methods, pixel-level
AUROC differed by at most 0.03 between strata and the AUROC ranking was
identical in both, as was the region-level Spearman ranking. Residual
misalignment is therefore unlikely to be the dominant driver of the
observed uncertainty--error relationship.
\subsection{Relationship Between Generation Uncertainty and Semantic Correctness}

In this section, we investigate whether generation uncertainty corresponds to semantic correctness. Specifically, we examine whether high uncertainty is associated with prediction errors, while low uncertainty corresponds to correct predictions. We systematically evaluate this relationship at three different scales: pixel-level, region-level, and image-level.


\begin{figure}[t]
    \centering
    \includegraphics[width=1.0\linewidth]{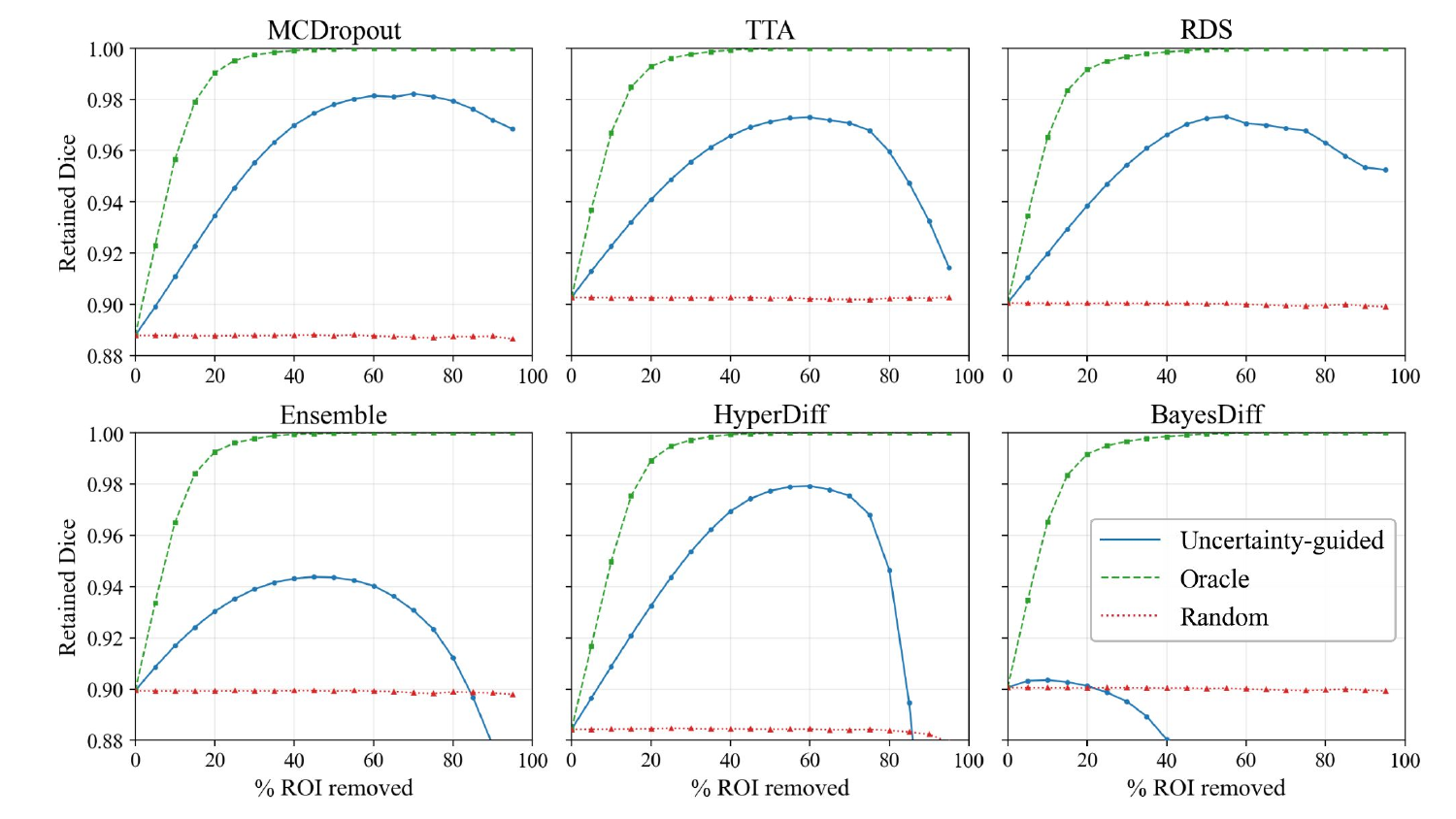}
    \caption{Pixel-wise sparsification curves for different uncertainty estimation methods. Pixels within the aortic region are gradually removed according to their uncertainty ranking, and the retained Dice is recomputed at each step. The uncertainty-guided curve (blue) is compared against the theoretical best (green) and random removal (red).}
    \label{fig:pixel_wise}
\end{figure}

\subsubsection{Pixel-Level Evaluation}

MCDropout's uncertainty corresponds most closely with pixel-level
semantic errors on AUSE, AUSC and AUROC (Table~\ref{tab:pixel_results}),
though HyperDiff and RDS achieve similar AUROC and the three form a
close top cluster; BayesDiff performs no better than random guessing
(AUROC 0.493) and diverges rapidly from the oracle in
Fig.~\ref{fig:pixel_wise}. Notably,
MCDropout and RDS reach this level as lightweight, retraining-free
methods, whereas HyperDiff requires training a hypernetwork ---
strong pixel-level correspondence is therefore achievable without
algorithmic changes or retraining.

The sparsification curves show additional information that the summary metrics do not.
MCDropout and HyperDiff both show the largest early gains in retained
Dice, confirming that their most-uncertain pixels are frequently
erroneous, but HyperDiff declines substantially at high sparsification
fractions while MCDropout tracks the oracle throughout. HyperDiff
assigns very low uncertainty to some individual error pixels (middle
panel, Fig.~\ref{fig:hyperdiffpixelregion}): it identifies highly
uncertain error pixels well, but assigns uncertainty less consistently
across the full ranking. This inconsistency is what its region-level
performance later recovers.

\begin{table}[t]
\centering
\caption{Pixel-level comparison of uncertainty estimation methods.
AUSE ($\downarrow$) and AUSC ($\uparrow$) summarise the pixel-wise
sparsification curve; AUROC ($\uparrow$) measures classification of
correct versus incorrect pixels. Best in bold.}
\label{tab:pixel_results}
\begin{tabular}{lccc}
\toprule
Method & AUSE$\downarrow$ & AUSC$\uparrow$ & AUROC$\uparrow$ \\
\midrule
TTA       & 0.037 & 0.904 & 0.787 ± 0.094 \\
Ensemble  & 0.063 & 0.878 & 0.689 ± 0.122\\
RDS       & 0.035 & 0.906 & 0.809 ± 0.094 \\
MCDropout & \textbf{0.028} & \textbf{0.911} & \textbf{0.830 ± 0.073} \\
HyperDiff & 0.049 & 0.890 & 0.815± 0.084 \\
BayesDiff & 0.170 & 0.771 & 0.493 ± 0.049 \\
\bottomrule
\end{tabular}
\end{table}



\subsubsection{Region-Level Evaluation}

\begin{figure}[t]
\centering
\includegraphics[width=1.0\linewidth]{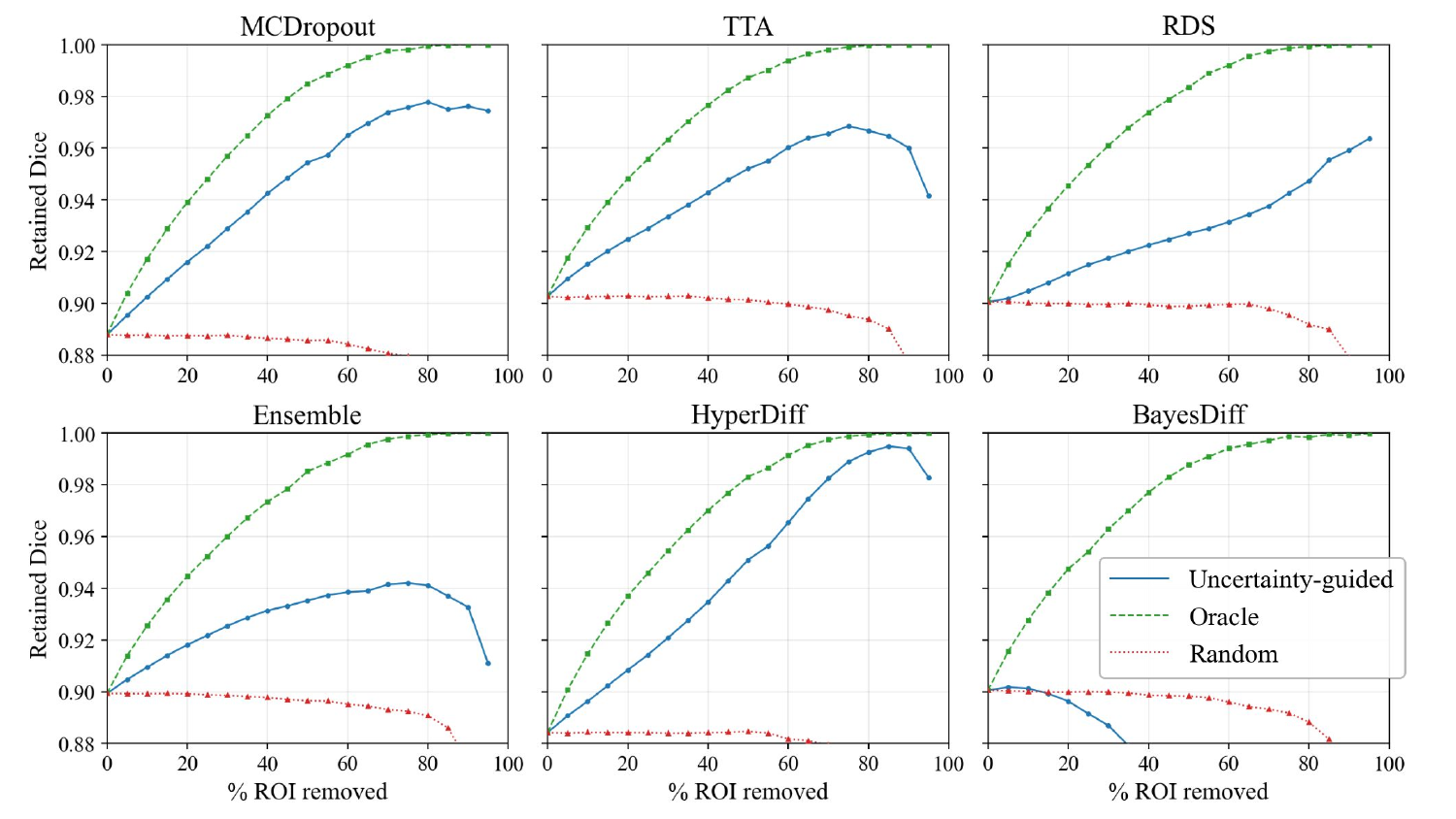}
\caption{Region-level sparsification curves. SLIC superpixels are
progressively removed in order of decreasing average uncertainty and
the retained Dice recomputed at each step, comparing each method's
uncertainty-guided curve (blue) against the oracle (green) and random
removal (red).}
\label{fig:localized_wise}
\end{figure}

\begin{figure}[t]
    \centering
    \includegraphics[width=1.0\linewidth]{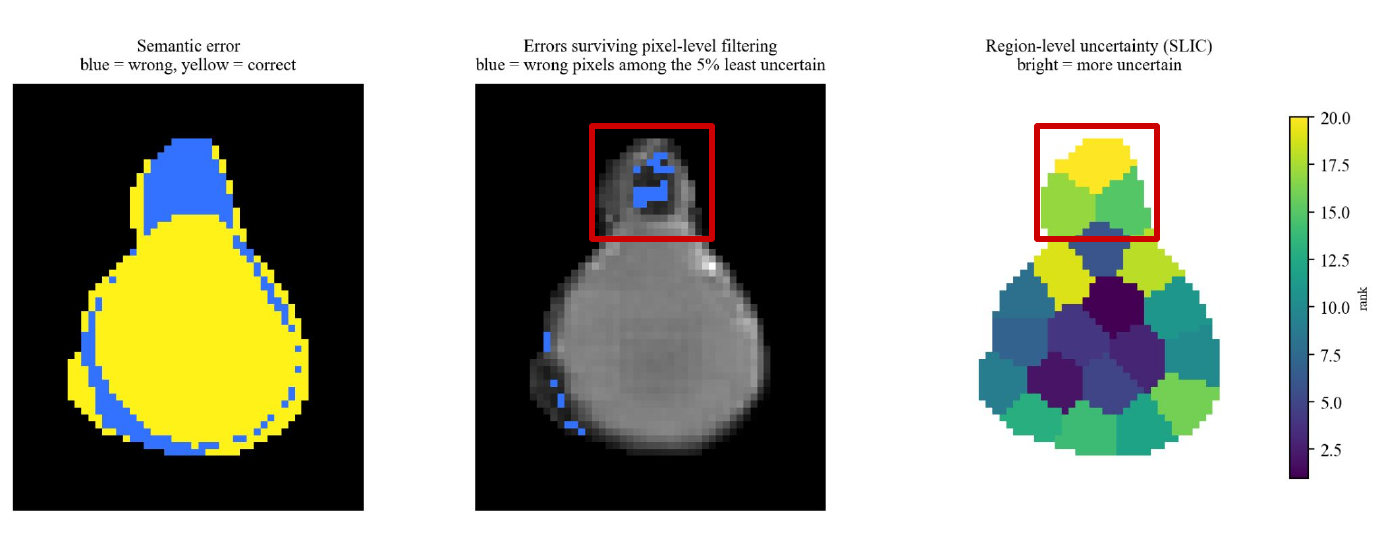}
    \caption{HyperDiff pixel-level under-flagging versus region-level uncertainty on a representative slice. }
    \label{fig:hyperdiffpixelregion}
\end{figure}

HyperDiff and MCDropout both outperform the other methods and are effectively tied across all three metrics (Fig.~\ref{fig:localized_wise}, Table~\ref{tab:localized_stats}), indicating that both localize error-containing regions effectively. BayesDiff is again weakest, with a negative correlation ($-0.24 \pm 0.28$) meaning its most uncertain regions tend to be less erroneous. The ranking differs from the pixel-level ranking, so the two scales capture distinct properties. HyperDiff's rise is the clearest case. It reaches the top cluster on localized sparsification despite its inconsistent pixel-level uncertainty because region-level scoring averages over each superpixel: the high uncertainty of neighbouring error pixels outweighs the under-flagged ones, so the region still scores high. Fig.~\ref{fig:hyperdiffpixelregion} shows this --- error pixels that survive pixel-level filtering (red box, middle panel) fall inside superpixels flagged as highly uncertain in the region-level map (right panel). Averaging smooths over exactly the inconsistency that limits HyperDiff at the pixel level.
\begin{table}[t]
\centering
\caption{Region-level comparison of uncertainty estimation methods.
AUSE ($\downarrow$) and AUSC ($\uparrow$) summarise the localized
sparsification curve; Spearman correlation ($\uparrow$) measures rank
agreement between region uncertainty and region semantic error.}
\label{tab:localized_stats}

\begin{tabular}{lccc}
\hline
Method & AUSE $\downarrow$ & AUSC $\uparrow$ & Spearman $\uparrow$\\ 
\hline
MCDropout & 0.023 & \textbf{0.901} & \textbf{0.65 ± 0.19} \\
TTA & 0.028 & 0.897 & 0.56 ± 0.23 \\
RDS & 0.042 & 0.881 & 0.60 ± 0.20 \\
Ensemble & 0.041 & 0.882 & 0.44 ± 0.29 \\
HyperDiff & \textbf{0.021} & 0.899 & 0.64 ± 0.16 \\
BayesDiff & 0.160 & 0.764 & -0.24 ± 0.28 \\
\hline
\end{tabular}

\end{table}

\begin{figure}[t]
\centering
\includegraphics[width=1.0\linewidth]{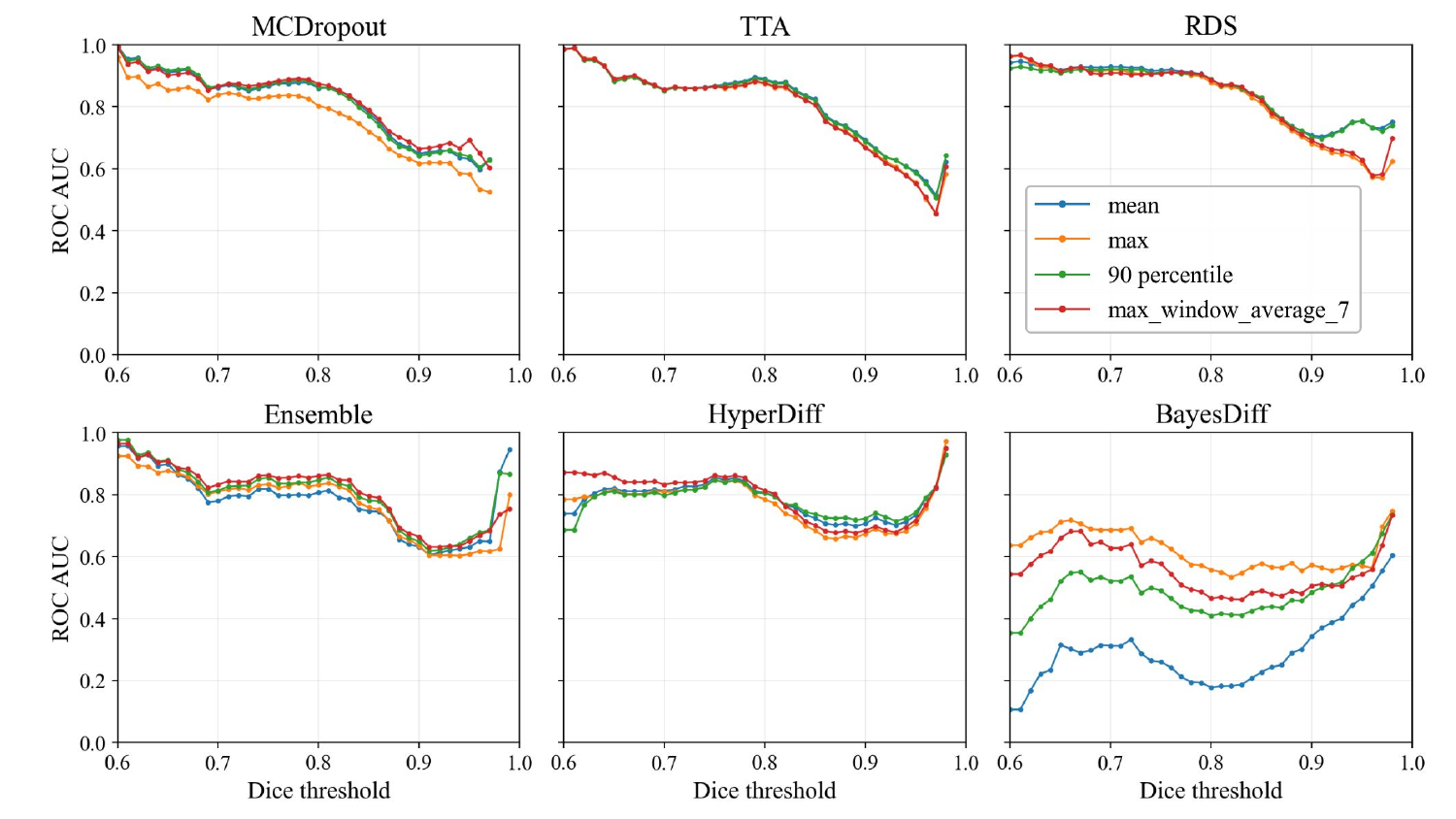}
\caption{Image-level failure detection. A sample is labelled failed
when its lumen Dice falls below a threshold; the threshold is swept
from 0.6 to 1.0 and AUROC reported at each.}
\label{fig:sample-classification}
\end{figure}
\subsubsection{Image-Level Evaluation}
Fig.~\ref{fig:sample-classification} reports AUROC as a function of the
Dice threshold that defines failure. MCDropout, TTA, RDS and Ensemble
exceed 0.9 for severe errors (Dice $<0.65$) and stay above 0.8 up to a
threshold of roughly 0.85. AUROC falls consistently as the threshold rises
from 0.6 to 0.9, indicating that uncertainty is most informative for
identifying severe failures.

\begin{figure}[t]
\centering
\includegraphics[width=1.0\linewidth]{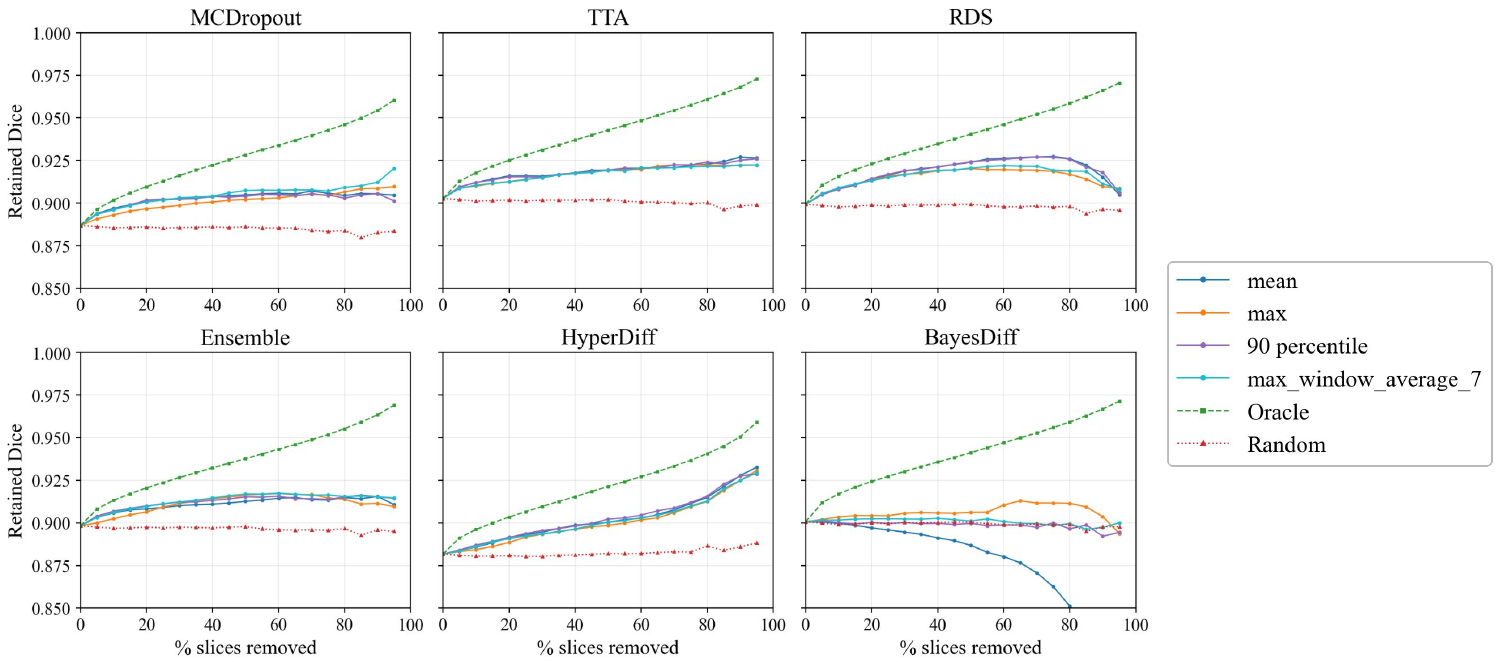}
\caption{Image-level sparsification curves. Images are progressively
removed in order of decreasing aggregated uncertainty and the average
Dice of the retained images recomputed at each step, compared against
the oracle (green) and random removal (red).}
\label{fig:sample-sparsification}
\end{figure}

Image-level sparsification assesses the ranking rather than the
detection (Fig.~\ref{fig:sample-sparsification}). HyperDiff follows the
oracle most closely and retains the highest Dice, reaching about 0.93
once 95\% of the most uncertain images are removed. MCDropout, TTA and
Ensemble are almost identical; RDS matches them early but declines at
high removal fractions, so its ranking is less reliable among the less
uncertain images; BayesDiff is worst throughout. All methods leave a
clear gap from the oracle, so none ranks images optimally and
uncertainty-guided filtering should be applied with care. The four
aggregation strategies give highly similar trends and leave the
relative ranking of methods unchanged, so these conclusions do not
depend on the aggregation choice.

\subsubsection{External Validation Under Distribution Shift}
\begin{figure}[t]
    \centering
    \includegraphics[width=1.0\linewidth]{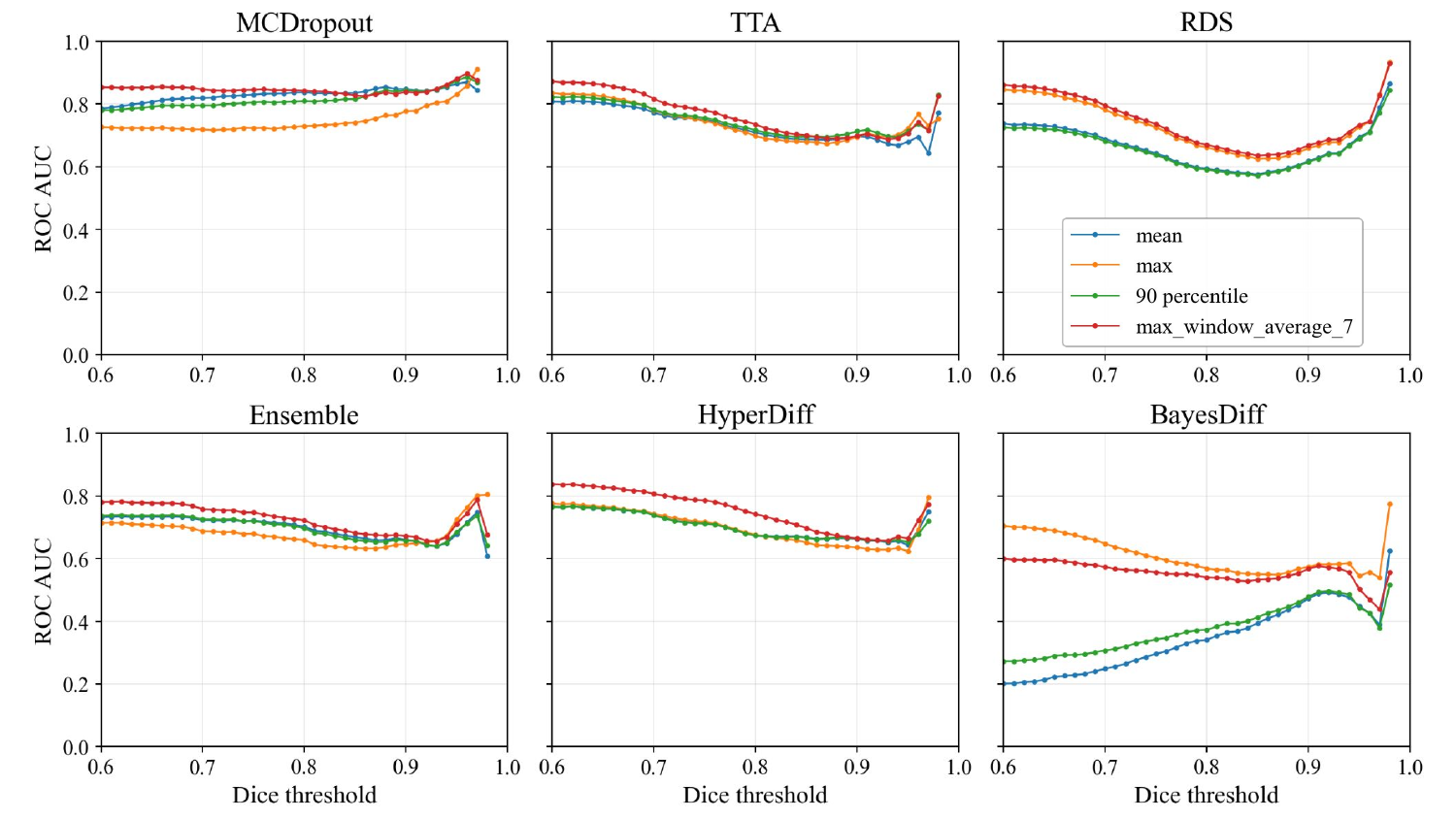}
    \caption{Image-level failure detection plots for different thresholds in AICT dataset. }
    \label{fig:sample-rejection-cmrad}
\end{figure}

We repeat the image-level failure detection analysis on the external
AICT dataset (Fig.~\ref{fig:sample-rejection-cmrad}). MCDropout is the
most consistent, exceeding 0.8 across most aggregation strategies and
nearly the entire threshold range; TTA, RDS and HyperDiff are
intermediate and largely preserve the internal ranking; BayesDiff
remains substantially weaker across most thresholds. The AUROC curves
are noticeably flatter than on the internal dataset, so failure
detection varies less across Dice thresholds under distribution shift.

\begin{figure}[t]
\centering
\includegraphics[width=1.0\linewidth]{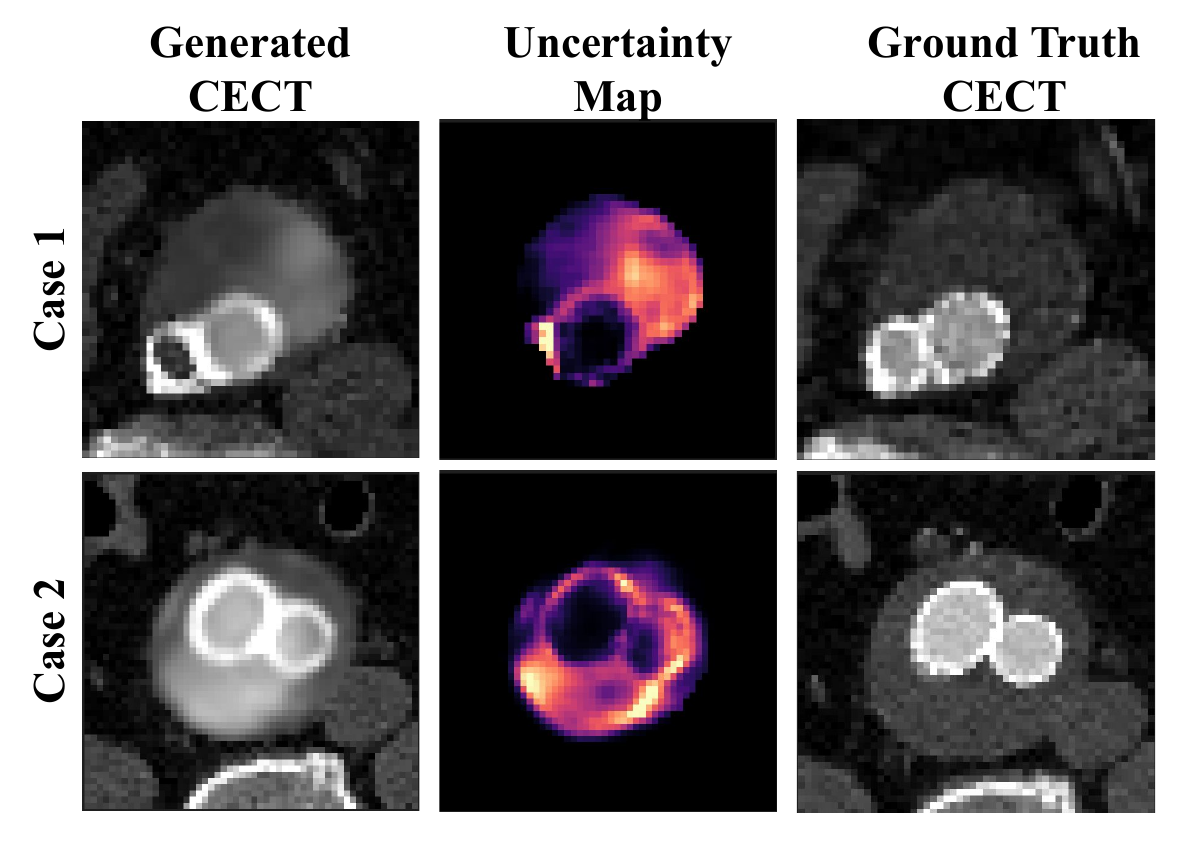}
\caption{Representative OOD cases containing vascular stents that were absent from the training dataset.}
\label{fig:external_visualization}
\end{figure}

Fig.~\ref{fig:external_visualization} shows representative external
cases. In both, uncertainty concentrates where the generated anatomy
deviates from the ground-truth CECT, corresponding to the generation
errors induced by the unseen structure, though the highest values are
not necessarily on the stent itself.

\subsubsection{Out-of-Distribution Detection}
We evaluate whether uncertainty can identify clinically relevant OOD
cases involving previously unseen vascular stents
(Table~\ref{tab:ood}). All methods except BayesDiff perform similarly, spanning 0.75--0.85
across aggregation strategies, with the highest value shared between
TTA, RDS and MCDropout depending on the strategy used. BayesDiff behaves differently from the other methods: mean
aggregation is weak (0.48) while maximum aggregation reaches 0.82, the
same pattern seen in the image-level analysis. As a reference, A KNN feature-space detector \cite{sun2022out} achieves only 0.43--0.44, substantially below every uncertainty-based method.

\begin{table}[t]
\centering
\caption{OOD stent detection AUROC ($\uparrow$) under four aggregation
strategies: mean, maximum, 90th percentile, and maximum local average
with window size 7.}
\label{tab:ood}

\begin{tabular}{lcccc}
\toprule
Method & Mean $\uparrow$ & Max $\uparrow$ & P90 $\uparrow$ & MWA$_7$ $\uparrow$ \\
\midrule
MCDropout & 0.77 & 0.83 & \textbf{0.81} & 0.82 \\
RDS       & 0.79 & 0.84 & 0.79 & \textbf{0.85} \\
HyperDiff & 0.75 & 0.78 & 0.75 & 0.81 \\
TTA       & \textbf{0.80} & \textbf{0.85} & \textbf{0.81} & 0.83 \\
BayesDiff & 0.48 & 0.82 & 0.53 & 0.72 \\
Ensemble  & 0.78 & 0.81 & 0.79 & 0.80 \\
\bottomrule
\end{tabular}
\end{table}



\section{Discussion}
Uncertainty provides meaningful information about semantic correctness in diffusion-based CECT synthesis, and can additionally identify clinically relevant out-of-distribution cases. Among the methods evaluated, MCDropout emerges as the method of choice: it performs strongly at every scale and under distribution shift, and requires no additional training when the base model already contains dropout, making it the practical default. In practice this makes MCDropout potentially deployable: a single uncertainty map,
produced without retraining, lets clinicians flag suspect pixels,
localize regions likely to contain errors, and reject unreliable
generations.

These differences follow from the source each method estimates.
BayesDiff places a posterior on the final convolutional layer alone,
treating all preceding layers as a fixed feature extractor, so its
uncertainty reflects one layer rather than the full denoiser --- a
plausible explanation for its consistently weak performance and its
negative region-level correlation. Ensemble and HyperDiff instead
capture variability across the entire network, and HyperDiff
concentrates it within spatially coherent regions, consistent with its
region-level strength. MCDropout produces many closely related
subnetworks whose disagreement varies smoothly across neighbouring
pixels, consistent with its stable pixel-level correspondence. TTA
reflects sensitivity to input changes, aligning with its OOD
performance, since unseen structures introduce input characteristics
the model is less robust to. RDS reflects output diversity across
initial noise realizations, and greater variability tends to indicate
lower-quality generations. The most appropriate method therefore
depends on the intended clinical application.

Computational requirements differ substantially
(Table~\ref{tab:methods}). TTA and RDS apply directly to any pretrained
model, and MCDropout is likewise training-free when the base model was
trained with dropout, whereas Ensemble and HyperDiff require additional
training. Since the training-free methods match or exceed the
retraining-based ones at every scale, deployment-time reliability
assessment need not carry a retraining cost --- relevant where
computational budget or energy footprint constrains clinical
deployment.

Several limitations should be acknowledged. The experiments were
conducted on a single application, NCCT to CECT synthesis of the
abdominal aorta, and generalizability to other organs, modalities and
generation problems remains to be validated empirically. We expect the
analysis to translate, since many medical image generation tasks
involve anatomically meaningful structures whose correctness can be
quantified through task-specific segmentations or structural
measurements, and the framework is independent of the particular
segmentation formulation used. In deployment, the estimates studied
here may help identify unreliable generations, localize likely semantic
errors, and alert clinicians to out-of-distribution structures or
artifacts. More broadly, these results suggest that uncertainty should be treated
not as a single quantity but as a multi-scale reliability signal, with
different mechanisms providing complementary information about a
generated image. The evaluation framework proposed here offers a
systematic way to study that signal in diffusion-based medical image
synthesis, and may support the development of more reliable and
deployable generative models for clinical use.

\section*{Acknowledgment}
We acknowledge the contribution from the Oxford Abdominal Aortic Aneurysm Study and the AICT Consortium (www.aict.ai) for the individual site’s equivalent approval of the AICT study protocol.

Specifically to the following: \textbf{Oxford Abdominal Aortic Aneurysm Study} (Ashok Handa, Pierfrancesco Lapolla-Lapolla, Anirudh Chandrashekar); \textbf{AICT Consortium} (Ewa Strauss, Grzegorz Oszkinis, Matt Bown, Michael Klontzas, Edwaldo Edner Joviliano, Saulo Da Silva Cordeiro, Lotte Pyfferoen, Kris Van De Moortele, Steven Van Wortswinkel, Filip Decker, John Kellas, Alex Pitman, David Chang, Juliette Raffor-Lareyre Fabien Lareyre, Stan Snijders, Daniel Feller, Ittai Dayan, Carlos Santin, Ernest Puyol, Anders Nordell, Erik Ranschaert)
Yuxuan Ou acknowledges the funding provided by the China Scholarship Council and University of Oxford through the CSC–PAG Oxford Scholarship.
Vicente Grau and Regent Lee are partially funded by NetZero AICT (EU New Horizon Grant), which has also funded data collection and curation for the AICT dataset.

\bibliographystyle{IEEEtran}
\bibliography{references}
\end{document}